%% file: main.tex
\documentclass[10pt,twocolumn,letterpaper]{article}

\usepackage{cvpr}              % To produce the CAMERA-READY version
\usepackage{multirow}
\usepackage{tablefootnote}
\usepackage{color, colortbl}

\input{preamble}

\definecolor{cvprblue}{rgb}{0.21,0.49,0.74}
\usepackage[pagebackref,breaklinks,colorlinks,citecolor=cvprblue]{hyperref}

\def\paperID{8125} % *** Enter the Paper ID here
\def\confName{CVPR}
\def\confYear{2024}

\title{FreGS: 3D Gaussian Splatting with Progressive Frequency Regularization}

\author{Jiahui Zhang\textsuperscript{\rm 1}
\quad Fangneng Zhan\textsuperscript{\rm 2} 
\quad Muyu Xu\textsuperscript{\rm 1} 
\quad Shijian Lu\textsuperscript{\rm 1} 
\quad Eric Xing\textsuperscript{\rm 3, 4} \\
\\[1mm]
{ $^1$Nanyang Technological University\quad$^2$Max Planck Institute for Informatics}\\[0.1mm]
{ $^3$Carnegie Mellon University\quad$^4$MBZUAI}
}

\begin{document}
\twocolumn[{
    \renewcommand\twocolumn[1][]{#1}
    \maketitle
    \begin{center}
        \captionsetup{type=figure}
        \includegraphics[width=1.\linewidth]{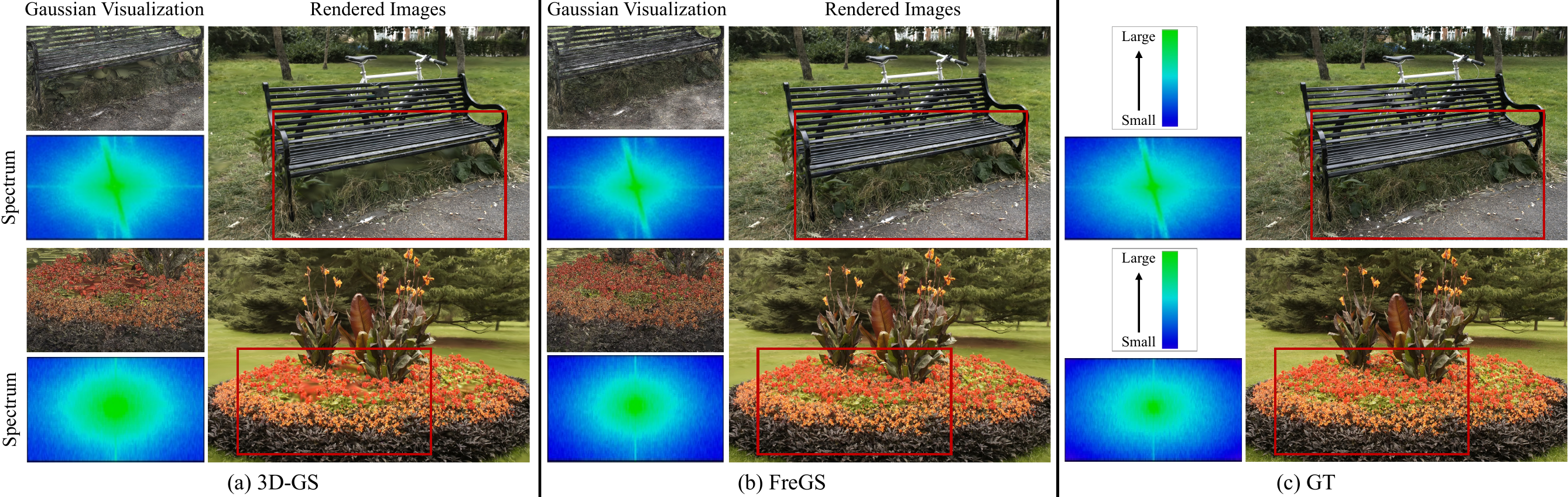}
        \captionof{figure}{
        The proposed FreGS mitigates the over-reconstruction of Gaussian densification and renders images with much less blur and artifact as compared with the 3D Gaussian splatting (3D-GS). For the two sample images from Mip-NeRF360~\cite{barron2022mip}, (a) and (b) show the \textit{Rendered Image} and the \textit{Gaussian Visualization} of the highlighted regions, as well as the \textit{Spectra} of over-reconstructed areas in the rendered image by 3D-GS and corresponding areas in FreGS. The \textit{Gaussian Visualization} shows how the learnt rasterized 3D Gaussians compose images (all Gaussians are rasterized with full opacity). The \textit{Spectra} are generated via image Fourier transformation, where the colour changes from blue to green as the spectrum amplitude changes from small to large.
        }
        \label{motivation}
    \end{center}

}]
\def\thefootnote{*}\footnotetext{Shijian Lu is the corresponding author.}
% \maketitle
\input{sec/0_abstract}   
\vspace{-1mm}
\input{sec/1_intro}
\input{sec/2_relatedwork}
\input{sec/3_method}
\input{sec/4_experiment.tex}
\input{sec/5_conclusion.tex}
{
    \newpage
    \small
    \bibliographystyle{ieeenat_fullname}
    \bibliography{main}
}


\end{document}

%% file: preamble.tex
\usepackage[dvipsnames]{xcolor}

%% file: sec/0_abstract.tex
\begin{abstract}
3D Gaussian splatting has achieved very impressive performance in real-time novel view synthesis. However, it often suffers from over-reconstruction during Gaussian densification where high-variance image regions are covered by a few large Gaussians only, leading to blur and artifacts in the rendered images. We design a progressive frequency regularization (FreGS) technique to tackle the over-reconstruction issue within the frequency space. Specifically, FreGS performs coarse-to-fine Gaussian densification by exploiting low-to-high frequency components that can be easily extracted with low-pass and high-pass filters in the Fourier space. By minimizing the discrepancy between the frequency spectrum of the rendered image and the corresponding ground truth, it achieves high-quality Gaussian densification and alleviates the over-reconstruction of Gaussian splatting effectively. Experiments over multiple widely adopted benchmarks (e.g., Mip-NeRF360, Tanks-and-Temples and Deep Blending) show that FreGS achieves superior novel view synthesis and outperforms the state-of-the-art consistently. 
\end{abstract}

%% file: sec/1_intro.tex
\section{Introduction}

Novel View Synthesis (NVS) has been a pivotal task in the realm of 3D computer vision which holds immense significance in various applications such as virtual reality, image editing, etc. It aims for generating images from arbitrary viewpoints of a scene, often necessitating precise modelling of the scene from multiple scene images. Leveraging implicit scene representation and differentiable volume rendering, NeRF~\cite{mildenhall2020nerf} and its extension \cite{barron2021mip, barron2022mip} have recently achieved remarkable progress in novel view synthesis. However, NeRF is inherently plagued by long training and rendering time. Though several NeRF variants \cite{muller2022instant, chen2022tensorf, garbin2021fastnerf, reiser2021kilonerf, fridovich2022plenoxels} speed up the training and rendering greatly, they often sacrifice the quality of rendered images notably, especially while handling high-resolution rendering.

As a compelling alternative to NeRF, 3D Gaussian splatting (3D-GS) \cite{kerbl20233d} has attracted increasing attention by offering superb training and inference speed while maintaining competitive rendering quality. By introducing anisotropic 3D Gaussians together with adaptive density control of Gaussian properties, 3D-GS can learn superb and explicit scene representations for novel view synthesis. It replaces the cumbersome volume rendering in NeRF by efficient splatting, which directly projects 3D Gaussians onto a 2D plane and ensures real-time rendering. However, 3D-GS often suffers from over-reconstruction \cite{kerbl20233d} during Gaussian densification, where high-variance image regions are covered by a few large Gaussians only which leads to clear deficiency in the learnt representations. The over-reconstruction can be clearly observed with blur and artifacts in the rendered 2D images as well as the discrepancy of frequency spectrum of the render images (by 3D-GS) and the corresponding ground truth as illustrated in Fig.~\ref{motivation}.

Based on the observation that the over-reconstruction manifests clearly by the discrepancy in frequency spectra, we design FreGS, an innovative 3D Gaussian splatting technique that addresses the over-reconstruction by regularizing the frequency signals in the Fourier space. FreGS introduces a novel frequency annealing technique to achieve progressive frequency regularization. Specifically, FreGS takes a coarse-to-fine Gaussian densification process by annealing the regularization progressively from low-frequency signals to high-frequency signals, based on the rationale that low-frequency and high-frequency signals usually encode large-scale (e.g., global patterns and structures which are easier to model) and small-scale features (e.g., local details which are harder to model), respectively. The progressive regularization strives to minimize the discrepancy of frequency spectra of the rendered image and the corresponding ground truth, which provides faithful guidance in the frequency space and complements the pixel-level L1 loss in the spatial space effectively. Extensive experiments show that FreGS mitigates the over-reconstruction and greatly improves Gaussian densification and novel view synthesis as illustrated in Fig.~\ref{motivation}.

The contributions of this work can be summarized in three aspects. 
\textit{First}, we propose FreGS, an innovative 3D Gaussian splatting framework that addresses the over-reconstruction issue via frequency regularization in the frequency space.
To the best of our knowledge, this is the first effort that tackles the over-reconstruction issue of 3D Gaussian splatting from a spectral perspective. 
\textit{Second}, we design a frequency annealing technique for progressive frequency regularization.
The annealing performs regularization from low-to-high frequency signals progressively, achieving faithful coarse-to-fine Gaussian densification.
\textit{Third}, experiments over multiple benchmarks show that FreGS achieves superior novel view synthesis and outperforms the 3D-GS consistently.

%% file: sec/2_relatedwork.tex
\section{Related Work}

\subsection{Neural Rendering for Novel View Synthesis}  

Novel view synthesis aims to generate new, unseen views of a scene or object from a set of existing images or viewpoints. 
When deep learning became popular in early days, CNNs were explored for novel view synthesis~\cite{hedman2018deep, riegler2020free, thies2019deferred}, e.g., they were adopted to predict blending weights for image rendering in~\cite{hedman2018deep}. Later, researchers exploit CNNs for volumetric ray-marching \cite{henzler2019escaping, sitzmann2019deepvoxels}.
For example, Sitzmann et al. propose Deepvoxels \cite{sitzmann2019deepvoxels} which builds a persistent 3D volumetric scene representation and then achieves rendering via volumetric ray-marching. 

Recently, neural radiance field (NeRF) \cite{mildenhall2020nerf} has been widely explored to achieve novel view synthesis via implicit scene representation and differentiable volume rendering. It exploits MLPs to model 3D scene representations from multi-view 2D images and can generate novel views with superb multi-view consistency. A number of NeRF variants have also been designed for handling various challenging conditions, such as dynamic scenes \cite{park2021nerfies, du2021neural, gao2021dynamic, guo2021ad, tretschk2021non}, free camera pose \cite{meng2021gnerf, lin2021barf, zhang2022vmrf, bian2023nope, zhang2023pose} and few shot \cite{chen2021mvsnerf, niemeyer2022regnerf, jain2021putting, yang2023freenerf}, antialiasing \cite{barron2022mip}. However, novel view synthesis with NeRF often comes at the expense of extremely long training and rendering times. Several studies \cite{muller2022instant, garbin2021fastnerf, reiser2021kilonerf, chen2022tensorf, fridovich2022plenoxels, hedman2021baking} attempt to reduce the training and rendering times. For example, KiloNeRF \cite{reiser2021kilonerf} speeds up NeRF rendering by utilizing thousands of tiny MLPs instead of one single large MLP. Chen et al. \cite{chen2022tensorf} leverage 4D tensor to represent full volume field and factorize the tensor into several compact low-rank tensor components for efficient radiance field reconstruction. Muller et al. propose InstantNGP \cite{muller2022instant} which introduces multi-resolution hash tables to achieve fast training and real-time rendering. 
However, most aforementioned work tends to sacrifice the quality of synthesized images especially while handling high-resolution rendering.

As a compelling alternative to NeRF, 3D Gaussian splatting \cite{kerbl20233d} introduces anisotropic 3D Gaussians and efficient differentiable splatting which enables high-quality explicit scene representation while maintaining efficient training and real-time rendering. 
However, the over-reconstruction of 3D Gaussians during Gaussian densification often introduces blur and artifacts in the rendered images. Our FreGS introduces progressive frequency regularization with frequency annealing to mitigate the over-reconstruction issue, enabling superior Gaussian densification and high-quality novel view synthesis.

\begin{figure*}[ht]
\begin{center}
\includegraphics[width=1\linewidth]{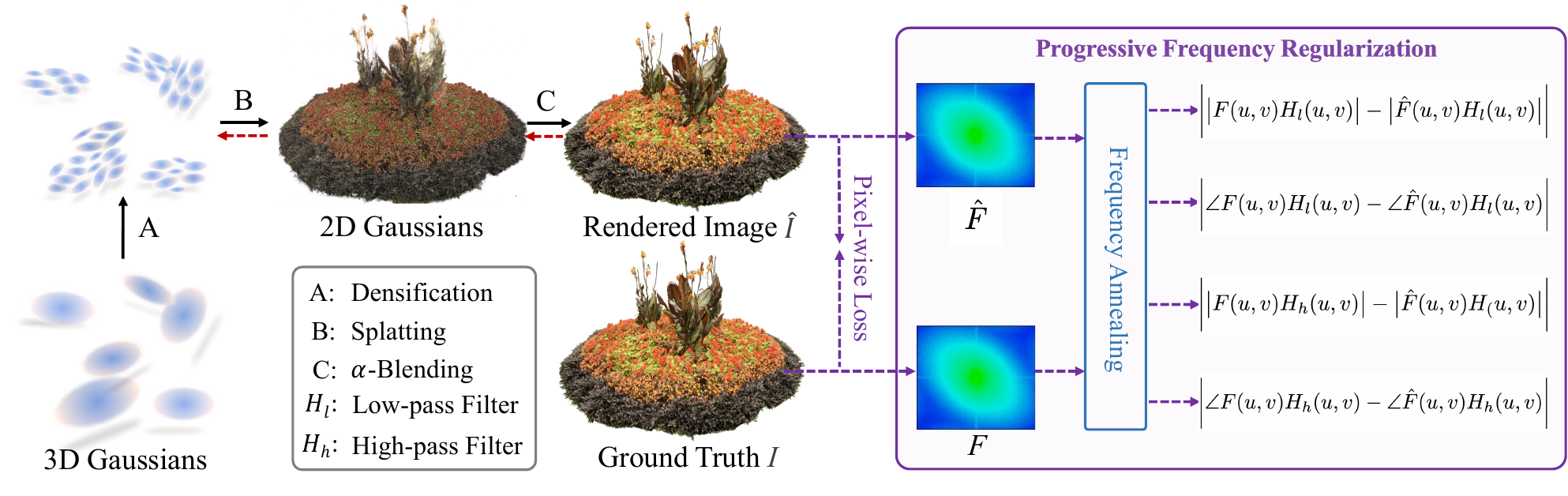}
\end{center}
\caption{
\textbf{Overview of the proposed FreGS.} 3D Gaussians are initialized by structure-from-motion. After splatting 3D Gaussians, we can obtain 2D Gaussians and then leverage standard $\alpha$-blending for rendering. Frequency spectra $\hat{F}$ and $F$ are generated by applying Fourier transform to rendered image $\hat{I}$ and ground truth $I$, respectively. Frequency regularization is achieved by regularizing discrepancies of amplitude $\big| F(u, v) \big|$ and phase$\angle F(u, v)$ in Fourier space. A novel frequency annealing technique is designed to achieve progressive frequency regularization. With low-pass filter $H_l$ and dynamic high-pass filter $H_h$, low-to-high frequency components are progressively leveraged to perform coarse-to-fine Gaussian densification. Note, the progressive frequency regularization is complementary to the pixel-wise loss between $\hat{I}$ and $I$. The red dashed line highlights the regularization process for Gaussian densification.
}
\label{overview}
\end{figure*}

\subsection{Frequency in Neural Rendering} 

NeRF has been widely explored in the frequency space. For example, \cite{mildenhall2020nerf} exploits sinusoidal functions of varying frequencies to encode inputs, overcoming the constraint of neural networks which often struggle to learn high-frequency information from low-dimensional inputs \cite{sitzmann2020implicit, tancik2020fourier, zhan2023general}. Several NeRF variants \cite{lin2021barf, park2021nerfies, wang2022hf, yang2023freenerf,xu2023wavenerf} also demonstrate the importance of learning in the frequency space under various challenging scenarios. For example, BARF \cite{lin2021barf} gradually increases frequency for learning NeRF without camera poses. WaveNeRF \cite{xu2023wavenerf} introduces wavelet frequency decomposition into multi-view stereo for achieving generalizable NeRF. Differently, FreGS boosts 3D Gaussian splatting in the frequency space, demonstrating that progressive frequency regularization with frequency annealing can lead more effective Gaussian densification and advanced novel view synthesis.

%% file: sec/3_method.tex
\section{Proposed Method}
We propose FreGS, a novel 3D Gaussian splatting with progressive frequency regularization which is the first to alleviate the over-reconstruction issue of 3D Gaussian splatting from frequency perspective. Fig.~\ref{overview} shows the overview of FreGS. The original 3D Gaussian splatting \cite{kerbl20233d} (3D-GS), including Gaussian densification, is briefly introduced in Sec.~\ref{3d-gs}. In Sec.~\ref{3_2}, we first reveal the reason for the effectiveness of frequency regularization in addressing the over-reconstruction issue and improving Gaussian densification. Then, we describe amplitude and phase discrepancies employed for frequency regularization within the Fourier space. To reduce the difficulty of Gaussian densification, we design frequency annealing technique (Sec.~\ref{3_3}) to achieve progressive frequency regularization, which can gradually exploit low-to-high frequency components to perform coarse-to-fine Gaussian densification.

\subsection{Preliminary}
\label{3d-gs}

\paragraph{3D Gaussian Splatting.} 3D-GS models scene representations explicitly with anisotropic 3D Gaussians and achieves real-time rendering by efficient differentiable splatting. Given a sparse point cloud generated by structure-from-motion \cite{hartley2003multiple, schonberger2016structure}, a set of 3D Gaussians is created, each of which is represented by a covariance matrix $\Sigma$, center position $p$, opacity $\alpha$ and spherical harmonics coefficients representing color $c$, where the covariance matrix $\Sigma$ is represented by scaling matrix and rotation matrix for differentiable optimization.

Gaussian densification aims to transform the initial sparse set of Gaussians to a more densely populated set, enhancing its ability to accurately represent the scene. It mainly focuses on two cases. The first is the regions with missing geometric features (corresponding to under-reconstruction) while the other is the large high-variance regions covered by a few large Gaussians only (corresponding to over-reconstruction). Both of these cases result in inadequate representation of regions within scenes. For under-reconstruction, Gaussians are densified by cloning the Gaussians, which increases both the total volume and the number of Gaussians. For over-reconstruction, Gaussian densification is achieved by dividing large Gaussians into multiple smaller Gaussians, which keeps the total volume but increases the number of Gaussians.

For rendering, 3D Gaussians are projected to a 2D plane by splatting. The rendering can then be achieved via the $\alpha$-blending. Specifically, the color $C$ of a pixel can be computed by blending $N$ ordered 2D Gaussians that overlap the pixel
, which can be formulated by:
\begin{equation}
C = \sum_{i \in N} c_i \alpha_i \prod_{j=1}^{i-1}(1-\alpha_j),
\end{equation}
where the color $c_i$ and $\alpha_i$ are calculated by multiplying the covariance matrix of $i$-th 2D Gaussian by per-point spherical harmonics coefficients and opacity, respectively. 

\subsection{Frequency Regularization}
\label{3_2}

\begin{figure}[t]
\begin{center}
\includegraphics[width=1\linewidth]{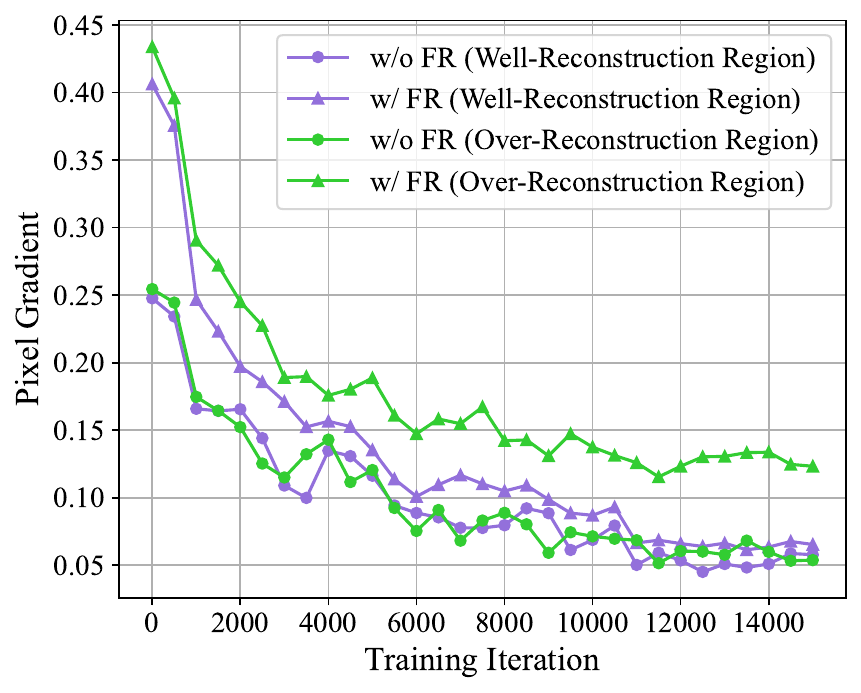}
\end{center}
\caption{
Average pixel gradients within \textbf{over-reconstruction regions} and \textbf{well-reconstruction regions} in scene `Bicycle’. The curve with circle (w/o frequency regularization (FR)) represents the method equivalent to 3D-GS \cite{kerbl20233d}, which utilizes pixel-wise L1 loss in the spatial domain only. As the Gaussian densification is terminated after the 15000$th$ iteration as in 3D-GS, we only show comparisons before the 15000$th$ iteration. It can be observed that the frequency regularization can increase the pixel gradient within \textit{over-reconstruction regions} significantly. Thus, compared with L1 loss, the frequency regularization shows superior capability in revealing the over-reconstruction region.
}
\label{curve}
\end{figure}

In this section, we first explore the reason why 3D-GS leads to over-reconstruction. We compute the average gradient of pixels within the \textit{over-reconstruction regions}, tracking its changes as training progresses. As Fig.~\ref{curve} shows, with a naive pixel-wise L1 loss, the average gradient could be quite small although the regions are not well reconstructed, which misleads the Gaussian densification. Specifically, the small pixel gradients are back-propagated to 2D splats for this pixel and the corresponding 3D Gaussians. As Gaussian densification is not applied to Gaussians with small gradients \cite{kerbl20233d}, these Gaussians cannot be densified through splitting into smaller Gaussians, leading to over-reconstruction. The consequence of over-reconstruction is an insufficient representation of regions, marked by deficiencies in 
both overall structure (low-frequency information) and details (high-frequency information). 
Compared with pixel space, the over-reconstruction region can be better revealed in frequency space by explicitly disentangling different frequency components.
Thus, it is intuitive to guide the Gaussian densification by explicitly applying regularization in frequency domain. Fig.~\ref{curve} shows that the average pixel gradient increases significantly with frequency regularization, demonstrating its effectiveness. We thus conclude that with frequency regularization, Gaussians can be adaptively densified in the over-reconstruction regions. In contrast, L1 loss cannot differentiate over-reconstructed regions from well-reconstructed ones, leading to many redundant Gaussians created in well-reconstructed regions.

Based on the above analysis, we design FreGS which aims to boost 3D Gaussian splatting from frequency perspective. Specifically, it alleviates over-reconstruction and improves Gaussian densification by minimizing the discrepancy between the frequency spectrum of rendered images and corresponding ground truth. Amplitude and phase, as two major elements of frequency, can capture different information of the image. Therefore, we achieve the frequency regularization by regularizing the amplitude and phase discrepancies between rendered images $\hat{I} \in \mathbb{R}^{H \times W \times C}$ and ground truth $I \in \mathbb{R}^{H \times W \times C}$ within Fourier space. 

Here, we detail the amplitude and phase discrepancies. We first convert $\hat{I} $ and $I$ to corresponding frequency representations $\hat{F}$ and $F$ by 2D discrete Fourier transform. Take $I$ as an example:
\begin{equation}
F(u, v) = \sum_{x=0}^{H-1} \sum_{y=0}^{W-1} I(x, y) \cdot e^{-i2\pi(u\frac{x}{H}+v\frac{y}{W})},
\label{fuerior}
\end{equation}
where $(x, y)$ and $(u, v)$ represent the coordinates in an image and its frequency spectrum, respectively. $I(x, y)$ and $F(u,v)$ denote the pixel value and complex frequency value, respectively. 
Then, $F(u,v)$ can be expressed in terms of amplitude $\big | F(u, v) \big |$ and phase $\angle F(u, v)$ as below:
\begin{equation}
\big | F(u, v) \big | = \sqrt{Re(u, v)^2 + Im(u,v)^2}
\end{equation}
\begin{equation}
\angle F(u, v) = arctan(\frac{Im(u,v)}{Re(u,v)}),
\end{equation}
where $Im(u, v)$ and $Re(u, v)$ represent the imaginary components and the real components of $F(u, v)$.

The amplitude and phase discrepancies (denoted as $d_a$ and $d_p$) between the rendered image $\hat{I}$ and the ground truth $I$ can be obtained with the Euclidean metric. Besides, we compute the amplitude and phase of all frequency components to assess the disparities accurately, which are then averaged to derive the final discrepancies as follows:
\begin{equation}
d_a = \frac{1}{\sqrt{HW}} \sum_{x=0}^{H-1} \sum_{y=0}^{W-1} \bigg| \big| F(u, v) \big| - \big| \hat{F}(u, v) \big| \bigg|
\label{amplitude}
\end{equation}
\begin{equation}
d_p = \frac{1}{\sqrt{HW}} \sum_{x=0}^{H-1} \sum_{y=0}^{W-1} \bigg| \angle F(u, v) - \angle \hat{F}(u, v) \bigg|,
\label{phase}
\end{equation}
where $F(u, v)$ and $\hat{F}(u, v)$ denote the complex frequency value of $I$ and $\hat{I}$, respectively.

\subsection{Frequency Annealing}
\label{3_3}

\begin{figure}[t]
\begin{center}
\includegraphics[width=1\linewidth]{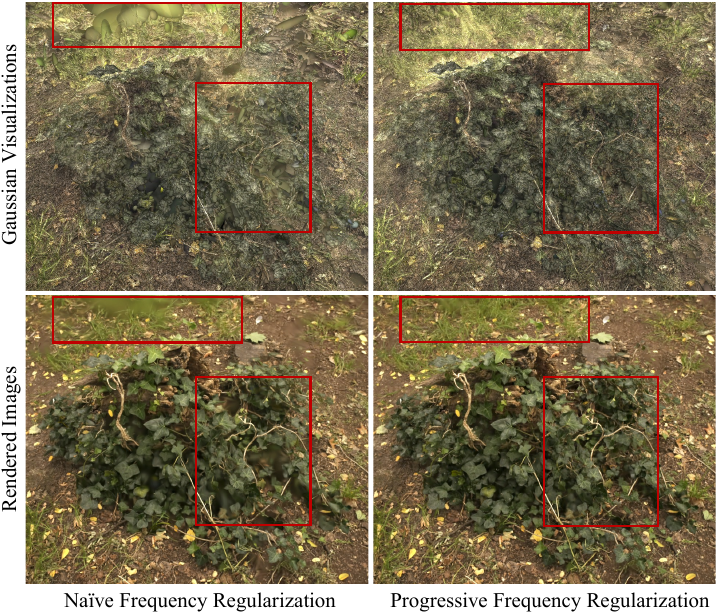}
\end{center}
\caption{
\textbf{The comparison of different frequency regularizations.} The naive frequency regularization directly employs amplitude and phase discrepancies without distinguishing between low and high frequency. The proposed progressive frequency regularization introduces frequency annealing technique to achieve low-to-high frequency regularization for coarse-to-fine Gaussian densification. It can be observed that the proposed progressive frequency regularization can achieve finer Gaussian densification and superior novel view synthesis. \textbf{Zoom in for best view.}
}
\label{fre_regularization}
\end{figure}

Though naively adopting the amplitude and phase discrepancies (without distinguishing between low and high frequency) as the frequency regularization can mitigate over-reconstruction to some extent, it still suffers from restricted Gaussian densification and significantly biases 3D Gaussian splatting towards undesirable artifacts (as shown in Fig.~\ref{fre_regularization}). As low and high frequency relates to large-scale features (e.g., global patterns and structures) and small-scale features (e.g., local details), respectively, we design frequency annealing technique to perform progressive frequency regularization, which gradually leverages low-to-high frequency to achieve coarse-to-fine Gaussian densification. With frequency annealing technique, superior Gaussian densification can be achieved as shown in Fig.~\ref{fre_regularization}. 

Specifically, to achieve frequency annealing, we incorporate the low-pass filter $H_l$ and dynamic high-pass filter $H_h$ in Fourier space to extract low and high frequency (denoted as $LF(u,v)$ and $HF(u,v)$), respectively. 
\begin{equation}
LF(u,v) = F(u, v) H_l(u, v)
\end{equation}
\begin{equation}
HF(u,v) = F(u, v) H_h(u, v).
\end{equation}
The corresponding amplitude and phase discrepancies for low and high frequency can then be formulated as follows:
\begin{equation}
d_{la} = \frac{1}{\sqrt{HW}} \sum_{x=0}^{H-1} \sum_{y=0}^{W-1} \bigg| \big| LF(u, v) \big| - \big| \hat{LF}(u, v) \big| \bigg|
\end{equation}
\begin{equation}
d_{lp} = \frac{1}{\sqrt{HW}} \sum_{x=0}^{H-1} \sum_{y=0}^{W-1} \bigg| \angle LF(u, v) - \angle \hat{LF}(u, v) \bigg|
\end{equation}
\begin{equation}
d_{ha} = \frac{1}{\sqrt{HW}} \sum_{x=0}^{H-1} \sum_{y=0}^{W-1} \bigg| \big| HF(u, v) \big| - \big| \hat{HF}(u, v) \big| \bigg|
\end{equation}
\begin{equation}
d_{hp} = \frac{1}{\sqrt{HW}} \sum_{x=0}^{H-1} \sum_{y=0}^{W-1} \bigg| \angle HF(u, v) - \angle \hat{HF}(u, v) \bigg|
\end{equation}
where $d_{la}$, $d_{lp}$, $d_{ha}$ and $d_{hp}$ represent low-frequency amplitude discrepancy, low-frequency phase discrepancy, dynamic high-frequency amplitude discrepancy and dynamic high-frequency phase discrepancy, respectively. 

\begin{figure*}[t]
\begin{center}
\includegraphics[width=1\linewidth]{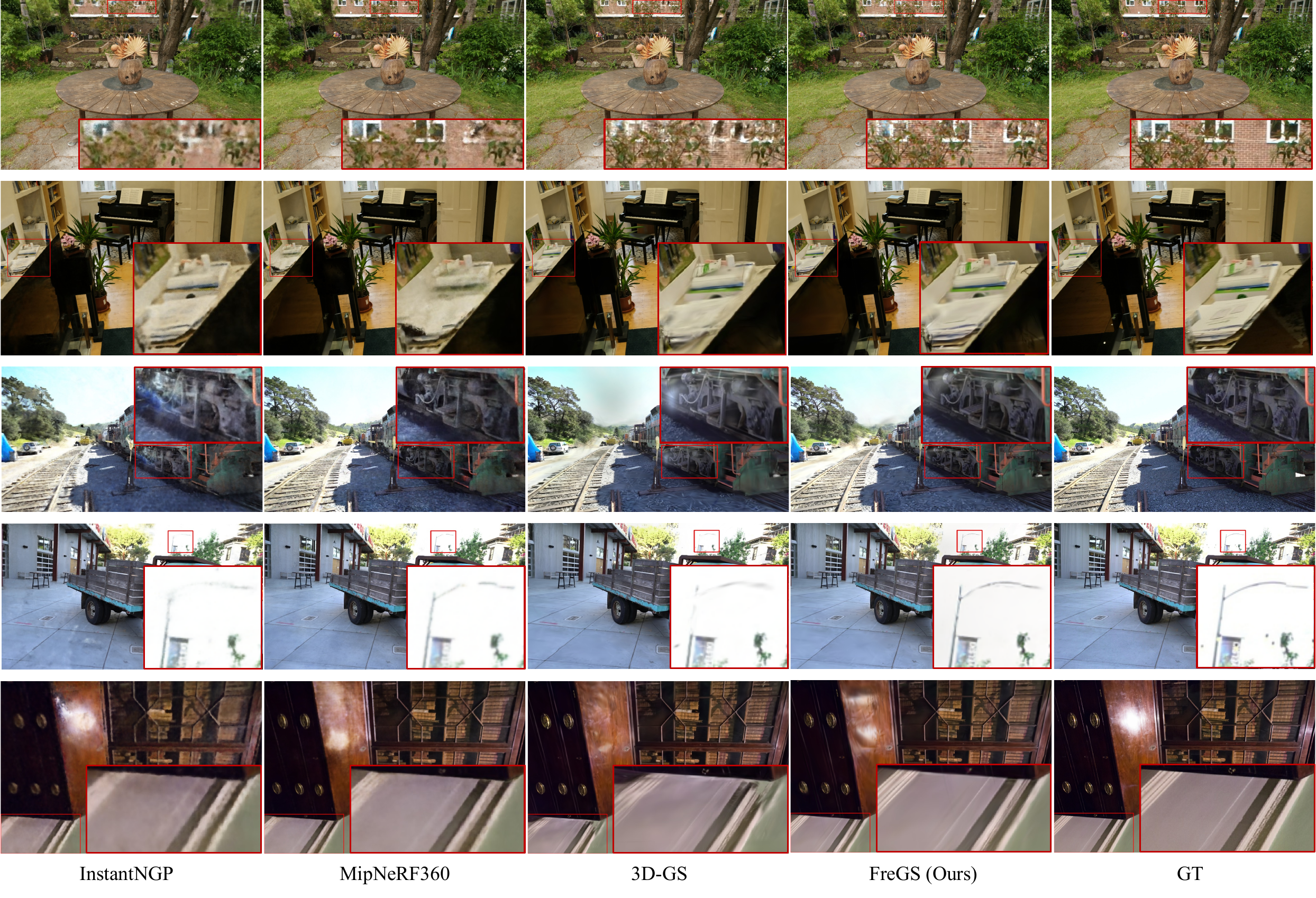}
\end{center}
\caption{
Qualitative comparisons of FreGS with three state-of-the-art methods in novel view synthesis. \textit{\textbf{\textcolor{red}{Note that} for fair comparison as well as trade-off balance between synthesis quality and memory consumption, we train FreGS with similar number of Gaussians as 3D-GS for these datasets (details in Sec.4.2). }} The comparisons are conducted over multiple indoor and outdoor scenes including `Garden' and `Room' from Mip-NeRF360, `Train' and `Truck' from Tank\&Temple, and `Drjohnson' from Deep Blending. `GT' denotes the ground-truth images. FreGS achieves superior image rendering with much less artifacts but more fine details.}
\label{qualitative_comparison}
\end{figure*}

For the progressive frequency regularization with the frequency annealing, we initiate it by regularizing the low-frequency discrepancies and then gradually integrate the high-frequency components as the training progresses. The gradual incorporation of high frequency can be achieved with the dynamic high-pass filter $H_h$, where the frequency band range $D_t$ allowed to pass at the $t$-th ($t\in[T_0, T]$) iteration can be expressed by:
\begin{equation}
D_0< D_t < \frac{(t-T_0)(D-D_0)}{T-T_0} + D_0,
\end{equation}
where $D_0$ and $D$ denote the maximum range allowed by the low-pass filter and the maximum range of frequency spectrum, respectively. Note that we take the center point $(H/2, W/2)$ as the coordinate origin. $t$, $T_0$ and $T$ represent the current iteration, the starting and end iterations of introducing high-frequency components, respectively. Regularization applied to low-to-high frequency results in coarse-to-fine Gaussian densification. The progressive frequency regularization $\mathcal{L}_f$ can be formulated as follows:
\begin{equation}
\mathcal{L}_f = \left\{
\begin{aligned}
    & w_l (d_{la} + d_{lp}), 0 < t \leq T_0 \\
    & w_l (d_{la} + d_{lp}) + w_h (d_{ha} + d_{hp}), t > T_0, 
\end{aligned}
\right.
\end{equation}
where $w_l$ and $w_h$ represent the training weights for low frequency and high frequency, respectively.

%% file: sec/4_experiment.tex
\section{Experiments}

\subsection{Datasets and Implementation Details}
\paragraph{Datasets} 
\label{dataset}
For training and testing, we follow the dataset setting of 3D-GS \cite{kerbl20233d} and conduct experiments on images of a total of 11 real scenes. Specifically, we evaluate FreGS on all nine scenes of Mip-NeRF360 dataset \cite{barron2022mip} and two scenes from the Tanks\&Temples dataset \cite{knapitsch2017tanks}. The selected scenes exhibit diverse styles, ranging from bounded indoor environments to unbounded outdoor ones. To divide the datasets into training and test sets, we follow 3D-GS and allocate every 8$th$ photo to the test set. The resolution of all involved images is the same as in 3D-GS as well.

\begin{table*}[t]
\definecolor{rred}{rgb}{1,0.6,0.6}
\definecolor{orange}{rgb}{1,0.8,0.6}
\definecolor{yellow}{rgb}{1,1,0.6}
\small
        \renewcommand\arraystretch{1.0}
	\renewcommand\tabcolsep{10pt}
		\begin{tabular}{l|ccc|ccc|ccc}
                \hline
                
			Datasets & \multicolumn{3}{c|}{Mip-NeRF360}  & \multicolumn{3}{c|}{Tanks\&Temples} & \multicolumn{3}{c}{Deep Blending}\\
                \hline
			Methods
			& SSIM$^\uparrow$   & PSNR$^\uparrow$    & LPIPS$^\downarrow$
			& SSIM$^\uparrow$   & PSNR$^\uparrow$    & LPIPS$^\downarrow$
			& SSIM$^\uparrow$   & PSNR$^\uparrow$    & LPIPS$^\downarrow$\\
			\hline 
			Plenoxels& 0.626 & 23.08 & 0.463 & 0.719 & 21.08 & 0.379 & 0.795 & 23.06 & 0.510\\
			INGP-Base& 0.671 & 25.30 & 0.371 & 0.723 & 21.72 & 0.330 & 0.797 & 23.62 & 0.423  \\ 
			INGP-Big& 0.699 & 25.59 & 0.331 & 0.745 & 21.92 & 0.305 & 0.817 & 24.96 & 0.390 \\ 
			Mip-NeRF360& \cellcolor{yellow} 0.792 & \cellcolor{orange} 27.69 & \cellcolor{yellow} 0.237 & \cellcolor{yellow} 0.759 & \cellcolor{yellow} 22.22 & \cellcolor{yellow} 0.257 & \cellcolor{yellow} 0.901 & \cellcolor{yellow} 29.40 & \cellcolor{yellow} 0.245\\
			3D-GS& \cellcolor{orange} 0.815 & \cellcolor{yellow} 27.21 & \cellcolor{orange} 0.214 & \cellcolor{orange} 0.841 & \cellcolor{orange} 23.14 & \cellcolor{orange} 0.183 & \cellcolor{orange} 0.903 & \cellcolor{orange} 29.41 & \cellcolor{orange} 0.243\\
                \hline
                FreGS(Ours)& \cellcolor{rred} 0.826 & \cellcolor{rred} 27.85 & \cellcolor{rred} 0.209 & \cellcolor{rred} 0.849 & \cellcolor{rred} 23.96 & \cellcolor{rred} 0.178 & \cellcolor{rred} 0.904 & \cellcolor{rred} 29.93 & \cellcolor{rred} 0.240 \\
                \hline
			
		\end{tabular}
	\caption{Quantitative comparisons on the dataset Mip-NeRF360, Tank\&Temple and Deep Blending. \textit{\textbf{\textcolor{red}{Note that} for fair comparison as well as trade-off balance between synthesis quality and memory consumption, we train FreGS with similar number of Gaussians as 3D-GS for these datasets (details in Sec.4.2). }} All methods are trained with the same training data. INGP-Base and INGP-Big refer to the InstantNGP \cite{muller2022instant} with a basic configuration and a slightly larger network \cite{muller2022instant}, respectively. \colorbox{rred}{Best score}, \colorbox{orange}{second best score} and \colorbox{yellow}{thrid best score} are in red, orange and yellow respectively.
	}
 \label{quantitative_result}
\end{table*}

\renewcommand\arraystretch{1.1}
\begin{table}[t]
\renewcommand\tabcolsep{1.7pt}
\small
\begin{center}
    
\begin{tabular}{l|ccc|ccc}
                \hline
                
			Datasets & \multicolumn{3}{c|}{Mip-NeRF360}  & \multicolumn{3}{c}{Tank\&Temple}\\
                \hline
			Methods
			&   PSNR$^\uparrow$   & SSIM$^\uparrow$  & LPIPS$^\downarrow$ 
			&   PSNR$^\uparrow$   & SSIM$^\uparrow$   & LPIPS$^\downarrow$   \\
			\hline 
			Base & 27.21 & 0.815 & 0.214 & 23.14 & 0.841 & 0.183 \\
                \hline
                Base+FR & 27.63 & 0.818 & 0.213 & 23.76 & 0.844 & 0.181 \\
                \hline
                Base+FR+FA & 27.85 & 0.826 & 0.209 & 23.96 & 0.849  & 0.178 \\
                \hline
			
		\end{tabular}
\end{center} 
\caption{
Ablation studies of the proposed FreGS on the datasets Mip-NeRF360 and Tank\&Temple.
The baseline \textit{Base} adopts pixel-level L1 loss and the D-SSIM term for 3D Gaussian splatting in spatial space. Our \textit{Base+FR} incorporates frequency regularization (FR) to address the over-reconstruction in the frequency space. The \textit{Base+FR+FA} (i.e., FreGS) further introduces our proposed frequency annealing technique (FA) to achieve progressive frequency regularization. \textit{\textbf{\textcolor{red}{Note that}} for fair comparison, we train \textit{Base+FR+FA} with similar number of Gaussians as \textit{Base} by increasing the gradient threshold. Besides, \textit{Base+FR} and \textit{Base+FR+FA} have the same gradient threshold.}
}
\label{ablation}

\end{table}

\paragraph{Implementation} 
For progressive frequency regularization, we initiate it with low-frequency amplitude and phase discrepancies and then extend the regularization to progressively encompass high-frequency amplitude and phase discrepancies for fine Gaussian densification. Note, the pixel-level L1 loss in the spatial space plus the D-SSIM term is used in the whole training process, which complements the proposed progressive frequency regularization in the frequency space. We stop the Gaussian densification after the 15000th iteration as in 3D-GS \cite{kerbl20233d}. Note that the frequency regularization terminates once Gaussian densification ends. For stable optimization, we start the optimization by working with an image resolution that is four times smaller than the original images as in 3D-GS. After 500 iterations, we increase the image resolution to the original size by upsampling. We adopt Adam optimizer \cite{kingma2014adam} to train the FreGS and use the Pytorch framework \cite{paszke2019pytorch} for implementation. For the rasterization, we keep the custom CUDA kernels used in 3D-GS.

\subsection{Comparisons with the State-of-the-Art}

We compare FreGS with 3D-GS \cite{kerbl20233d} as well as other four NeRF-based methods \cite{fridovich2022plenoxels, muller2022instant, barron2022mip} over various scenes in datasets Mip-NeRF360 and Tank\&Temple. \textbf{For fair comparison as well as the trade-off balance between memory and performance, we train FreGS with a similar number of Gaussians as 3D-GS for these datasets, which is achieved by introducing progressive frequency regularization while increasing the gradient threshold.} All compared methods are trained with the same training data and hardware. Table~\ref{quantitative_result} shows experimental results over the same test images as described in Section~\ref{dataset}. We can observe that FreGS outperforms the state-of-the-art 3D-GS consistently in PSNR, SSIM and LPIPS across all real scenes. The superior performance is largely attributed to our proposed progressive frequency regularization which alleviates the over-reconstruction issue of Gaussians and improves the Gaussian densification effectively. In addition, FreGS surpasses Mip-NeRF360, INGP-Base, INGP-Big, and Plenoxels by significant margins in terms of the image rendering quality. As Fig.~\ref{qualitative_comparison} shows, FreGS achieves superior novel view synthesis with less artifacts and finer details. 

\begin{figure*}[t]
\begin{center}
\includegraphics[width=1\linewidth]{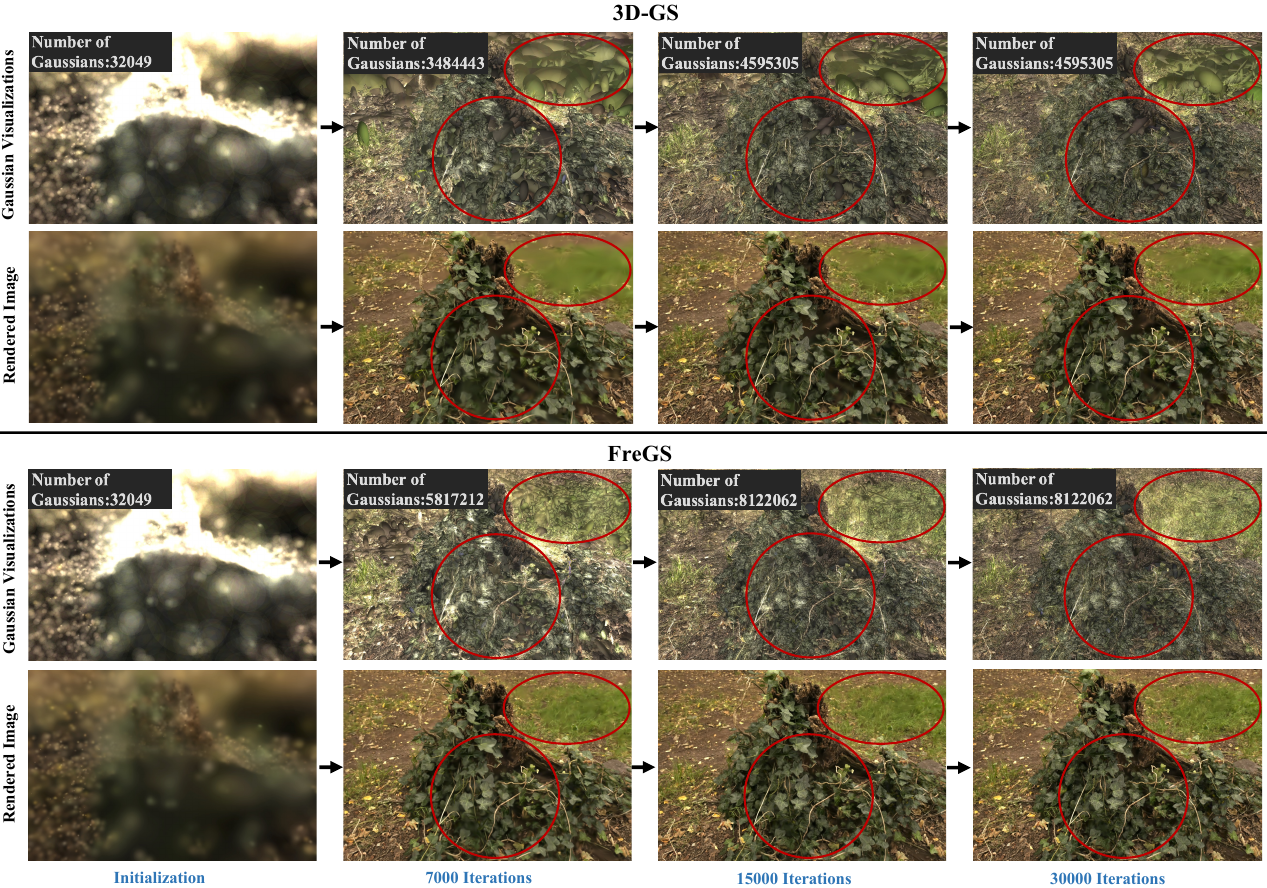}
\end{center}
\caption{Visualization of Gaussian densification and rendered images along the training process. The proposed FreGS improves the Gaussian densification for over-reconstruction regions progressively and the rendered images also improve progressively with less artifacts and finer details. We visualize both 3D-GS and FreGS at 7000, 15000, and the final 30000 training iterations for comparison. Note that the number of Gaussians does not change any more after the 15,000$th$ iteration due to the termination of Gaussian densification. 
}
\label{visualization}
\end{figure*}

\subsection{Ablation Studies}

We conduct ablation experiments to examine the proposed frequency regularization that mitigates the over-reconstruction in the frequency space, as well as the frequency annealing that achieves progressive frequency regularization. More details are described in the following two subsections.

\paragraph{Frequency Regularization} We first examine how our proposed frequency regularization affects PSNR, SSIM and LPIPS in Gaussian splatting. First, we train a baseline model \textit{Base} that performs Gaussian splatting with a L1 loss in the spatial space. On top of the \textit{Base}, we train a model \textit{Base+FR} that incorporates our proposed frequency regularization in the frequency space. As Table \ref{ablation} shows, \textit{Base+FR} outperforms the \textit{Base} clearly in PSNR, SSIM and LPIPS, indicating the critical role of frequency regularization in 3D Gaussian splatting and novel view synthesis from a frequency perspective.

\paragraph{Frequency Annealing} We then examine how our proposed frequency annealing affects Gaussian splatting. For that, we train a model \textit{Base+FR+FA} (i.e., FreGS) that incorporates the frequency annealing on top of \textit{Base+FR} for progressive frequency regularization. As Table \ref{ablation} shows, \textit{Base+FR+FA} outperforms \textit{Base+FR} clearly in novel view synthesis, demonstrating the effectiveness of the frequency annealing on progressive frequency regularization.

\subsection{Visualizations}

We visualize the Gaussian densification along the training process. As Fig.~\ref{visualization} shows, our FreGS 
produces more Gaussians and obtains clearly better Gaussian densification for over-reconstruction regions and superior novel view synthesis. Note that the number of Gaussians does not change any more after the 15,000$th$ iteration due to the termination of Gaussian densification. The visualization verifies that the proposed progressive frequency regularization mitigates the over-reconstruction of Gaussians and improves Gaussian densification effectively.

%% file: sec/5_conclusion.tex
\section{Conclusion}

This paper presents FreGS, an innovative 3D Gaussian splatting that explores progressive frequency regularization to boost 3D Gaussian splatting from a frequency perspective. Specifically, we design a frequency annealing technique for progressive frequency regularization, which performs coarse-to-fine Gaussian densification by progressively leveraging low-to-high frequency components that can be easily extracted with low-pass and high-pass filters in Fourier space. By minimizing the discrepancy between the frequency spectrum of rendered images and the corresponding ground truth, FreGS mitigates the over-reconstruction issue and achieves superior Gaussian densification. Experiments over multiple widely adopted indoor and outdoor scenes show that FreGS achieves superior novel view synthesis and outperforms the state-of-the-art consistently.

\section{Acknowledgements}

This project is funded by the Ministry of Education Singapore, under the Tier-2 project scheme with a project number MOE-T2EP20220-0003.

\clearpage